\documentclass[10pt,twocolumn,letterpaper]{article}

\usepackage{cvpr}              

\usepackage{graphicx}
\usepackage{amsmath}
\usepackage{amssymb}
\usepackage{booktabs}

\usepackage{multirow}
\usepackage{kotex}
\usepackage{pifont}
\newcommand{\xmark}{\ding{55}}
\usepackage{soul}

\usepackage[capitalize]{cleveref}
\crefname{section}{Sec.}{Secs.}
\Crefname{section}{Section}{Sections}
\Crefname{table}{Table}{Tables}
\crefname{table}{Tab.}{Tabs.}
\Crefname{figure}{Figure}{Figures}
\crefname{figure}{Fig.}{Figs.}

\def\cvprPaperID{8634} 
\def\confName{CVPR}
\def\confYear{2023}

\begin{document}

\title{Object-Centric Multi-Task Learning for Human Instances}

\author{Hyeongseok Son, Sangil Jung, Solae Lee, Seongeun Kim, Seung-In Park, ByungIn Yoo\\
Samsung Advanced Institute of Technology\\
{\tt\small \{hs1.son, sang-il.jung, solae913.lee, se91.kim, si14.park, byungin.yoo\}@samsung.com}
}
\maketitle

\newcommand{\change}[1]{{\color{red}#1}}
\newcommand{\son}[1]{{\textcolor{magenta}{hyeongseok: #1}}}
\newcommand{\park}[1]{{\textcolor{cyan}{seung-in: #1}}}
\newcommand{\sek}[1]{{\textcolor{blue}{\footnotesize seongeun: #1 \normalsize}}}
\newcommand{\il}[1]{{\color{red}#1}}
\newcommand{\sol}[1]{{\textcolor{yellow}{solae: #1}}}
\newcommand{\biyoo}[1]{{\textcolor{green}{\footnotesize bi: #1}}}

\newcommand{\SAMLOC}{\mathcal{S}}
\newcommand{\LEK}{\mathcal{K}}
\newcommand{\JNT}{\mathcal{J}}
\newcommand{\POSQ}{\mathcal{P}}
\newcommand{\CONQ}{\mathcal{E}}
\newcommand{\LOSS}{\mathcal{L}}

\newcommand{\noff}{N_p}
\newcommand{\nhead}{N_h}
\newcommand{\nscl}{N_s}

\newcommand{\REAL}{\mathbb{R}}
\newcommand{\PTS}{\mathbf{p}}

\begin{abstract}
 Human is one of the most essential classes in visual recognition tasks such as detection, segmentation, and pose estimation. Although much effort has been put into individual tasks, multi-task learning for these three tasks has been rarely studied. In this paper, we explore a compact multi-task network architecture that maximally shares the parameters of the multiple tasks via object-centric learning. To this end, we propose a novel query design to encode the human instance information effectively, called human-centric query (HCQ). HCQ enables for the query to learn explicit and structural information of human as well such as keypoints. Besides, we utilize HCQ in prediction heads of the target tasks directly and also interweave HCQ with the deformable attention in Transformer decoders to exploit a well-learned object-centric representation. Experimental results show that the proposed multi-task network achieves comparable accuracy to state-of-the-art task-specific models in human detection, segmentation, and pose estimation task, while it consumes less computational costs.
\end{abstract} 

\section{Introduction}
\label{sec:intro}

Core tasks of visual recognition are detection, segmentation, and pose estimation in scenes for the various vision applications such as video surveillance and human computer interaction.
Instead of training an individual model for each task, multi-task learning strategy using an unified architecture is beneficial in terms of cost-efficiency and inter-task synergy. In this paper, we focus on designing an effective unified architecture for the human-related multi-tasks.

Recently, object-centric  learning~\cite{Locatello01,carion2020detr} was introduced benefiting from Transformer architecture. Object-centric learning enables to encode per-instance representations by mapping queries in Transformer to corresponding object instances in an image. 
Due to the per-instance representation, object-centric learning is well-suited with human instance-level recognition tasks.
Although the object-centric learning has been applied in some studies on multi-task learning for detection and segmentation~\cite{li2022mask}, few attempts have been made to apply the concept to human recognition problems. 

Furthermore, unlike general object recognition, human recognition needs to estimate poses that imply the structural information of the human instances. Thus, a na\"ive introduction of object-centric learning would lead to performance degradation because different types of information are implicitly encoded in a mixed way, preventing each query from having tailored information for each task. 

\begin{figure*}[t]
\begin{center}
\includegraphics [width=1.0\linewidth] {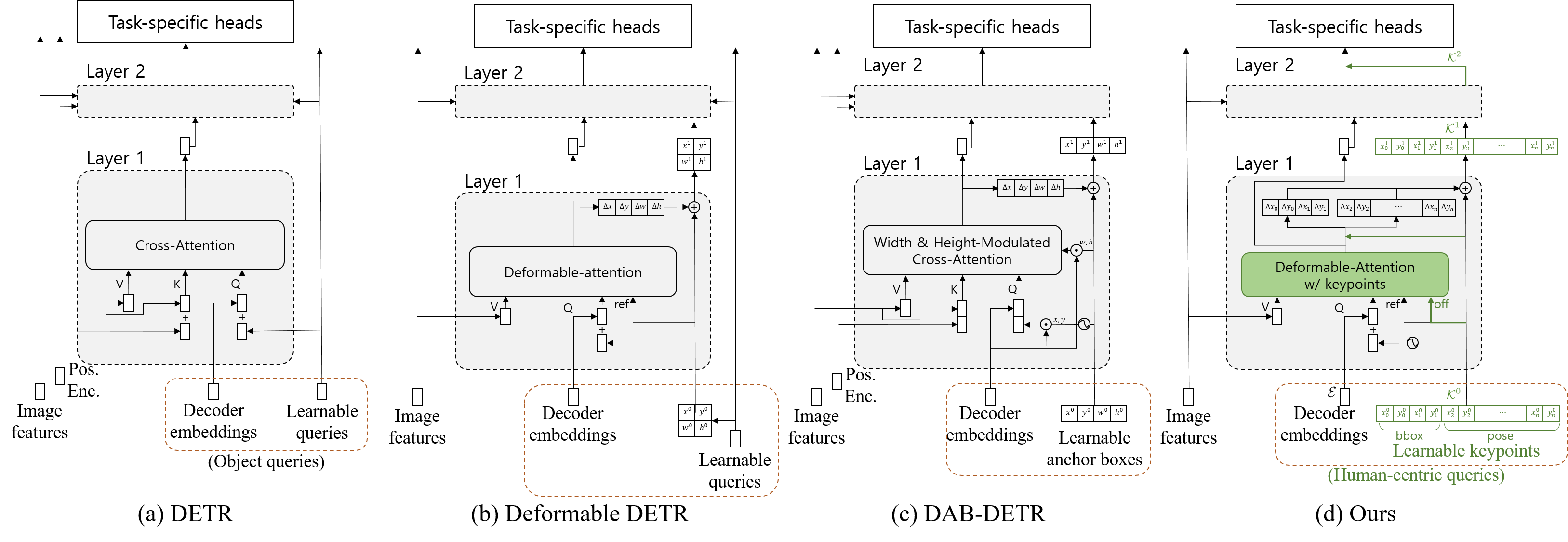}
\end{center}
\vspace{-12pt}
  \caption{Query design comparison of DETR~\cite{carion2020detr}, Deformable DETR~\cite{zhu2021deform}, DAB-DETR~\cite{liu2022dab} and ours.
  For clarity, we visualize a cross-attention function in Transformer decoders and highlight our main differences in green color.
  In terms of a query design, Deformable DETR and DAB-DETR embed more explicit positional information of an object to a learnable query. Our human-centric query additionally represents the structural information of an object as the form of keypoints. This enables our human-centric query and decoder embedding to carry various information for multiple tasks effectively.
  Besides, distinct from all the other methods, these learnable keypoints are fed in task-specific heads directly, and give pre-computed high-level positional and structural information to them. Then, each task performs better by considering such information together.
  In addition, the learnable keypoints are directly used in deformable attentions as the sampling locations of the attentions, which boosts the performance of all target tasks by exploiting learnable keypoints.
}
\label{fig:query}
\vspace{-8pt}
\end{figure*}

To tackle this problem, we propose a novel human-centric query (HCQ) design to extract more representative information of human instances. We segregate the structural information from the \textit{decoder embedding} and represent it in explicit forms called \textit{learnable keypoints} (\cref{fig:query}d). The learnable keypoints contain the bounding box and the body joints of a human while the decoder embedding encodes the general representation of an instance. This separation enables different pieces of information of a human instance to be encoded in a decoupled way, allowing multiple target tasks to effectively utilize useful information from the query for their own purposes.

Thanks to the decoupled design of HCQ, only the light-weight prediction heads are needed because the information is already disentangled enough for each task. In addition, using both information in the prediction heads further improve the performance because it gives extra information that the other query does not have. For example, human pose can help segmentation~\cite{zhang2019pose2seg} and \textit{vice versa}. Even though some previous methods (e.g., DAB-DETR~\cite{liu2022dab}) represent coarse information explicitly such as bounding boxes, but it does not give enough information to help other tasks (e.g., segmentation). Thus, most previous methods feed only decoder embeddings to the prediction heads to perform target tasks even though they have explicit box coordinates. 

Furthermore, our learnable keypoints can be interweaved with deformable cross-attention seamlessly. The deforamable attention~\cite{zhu2021deform} predicts the sample locations from the decoder embedding directly instead of computing similarity between the query and the image features. We can just apply our learnable keypoints to the deformable cross-attention as the sampling locations. 
It reduces computation for predicting the sampling locations and also the high-quality keypoints give extra structural information to be attended compared to the direct prediction from the decoder embeddings.

We demonstrate our method on the COCO dataset~\cite{Lin2014microsoft} with various ablation studies. To our best knowledge, our method is the first Transformer-based unified architecture for human-related multi-tasks, and achieves the comparable accuracy with the state-of-the-art task-specific models while having compact design.

To summarize, our contributions are as follows:
\begin{itemize}
    \item We present a unified network architecture performing multiple human instance-level vision tasks with minimal task overheads. 
    \item We propose a new object-centric query design to effectively represent the structural information of human instances. With the object-centric query, our model can leverage the disentangled high-level information directly in the prediction head. 
    In addition to that, the query design allows to interweave the structural information with deformation attention seamlessly. 
    \item We experimentally show that our method achieves the comparable accuracy with the state-of-the-art task-specific models while having cost-efficient architecture. 
\end{itemize}

\section{Related Work}
\label{sec:related}

\begin{figure*}[t]
\begin{center}
\includegraphics [width=0.90\linewidth] {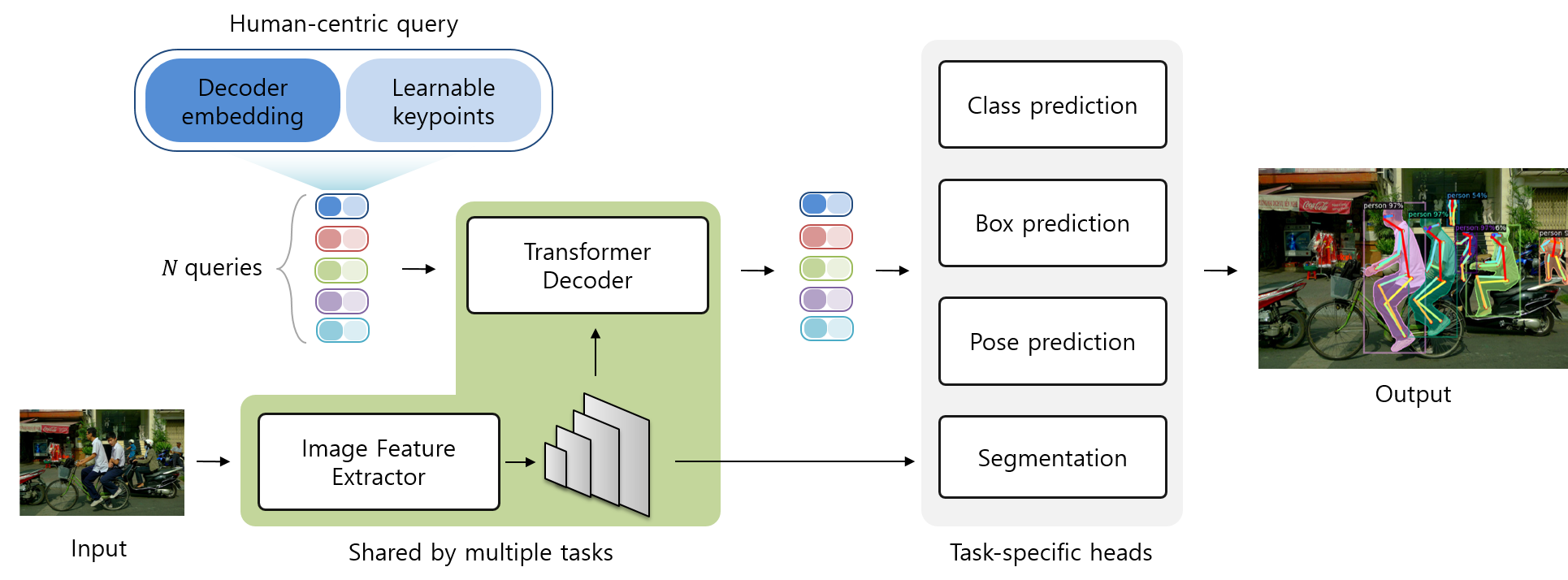}
\end{center}
\vspace{-12pt}
  \caption{Our unified network architecture for various instance-level vision tasks; human detection, segmentation and pose estimation.
}
\label{fig:overall}
\vspace{-8pt}
\end{figure*}

\paragraph{Object-centric learning}
The concept of object-centric representation learning is presented in Locatello \etal~\cite{Locatello01} where they focus on learning object representations in order to segregate information in an image according to the individual object entities. A set of object representations, called slots, are mapped to object instances in an image using slot attention. 
With the help of set prediction formulation, DETR~\cite{carion2020detr} proposed to learn object-centric representations with the Transformer architecture for object detection. They used a fixed-size set of learnable queries, called object queries, to infer the relations of the objects and the image feature (\cref{fig:query}(a)). 

\paragraph{Object query design}
Many follow-up approaches~\cite{meng2021cond,zhu2021deform,liu2022dab,wang2022anchor,li2022dn} have tackled the object detection task based on DETR.
Among them, we focus representative methods in perspective of object query designs.
In Deformable DETR~\cite{zhu2021deform}, queries are formulated as 2D reference points simply obtained by learned linear projection (\cref{fig:query}(b)). By leveraging 2D reference points as anchor points, it allows faster convergence and better performance in object detection. DAB-DETR~\cite{liu2022dab} replaced the query formulation with dynamic anchor boxes containing both position and size information (\cref{fig:query}(c)), hence each query conveys stronger positional priors to let decoders focus on a regions of interest. 
Distinct from the previous works, our learnable query explicitly and compactly carries the structural information of an object as the form of keypoints (\cref{fig:query}(d)). Furthermore, we present effective ways to exploit the high-level information learned in the query for performing deformable attention and task-specific heads.

\paragraph{Object-centric multi-task learning}
Object-centric representation has been applied to general object multi-task learning of detection and segmentation. 
DETR~\cite{carion2020detr} is an object detector benefiting from the object queries communicating with image features, can be easily adapted to segmentation task by attaching mask heads. 
Recently proposed Mask-DINO~\cite{li2022mask} further improved performance of the target tasks simultaneously by utilizing anchor box-guided cross-attention and denoising training scheme. In the segmentation point of view, three segmentation tasks including semantic segmentation, instance segmentation and panoptic segmentation~\cite{kirillov2019panoptic} can be regarded as different tasks, and some work~\cite{zhang2021k,cheng2022mask2former} conducted all these tasks in an unified architecture. Mask2Former~\etal~\cite{cheng2021maskformer,cheng2022mask2former} attended only the masked regions in cross-attention for fast convergence and K-net~\cite{zhang2021k} introduced learnable kernels with update strategy where each kernel is in charge of each mask.


\paragraph{Multi-task learning for human instances}
Human instance segmentation and pose estimation~\cite{papandreou2018personlab, zhang2019pose2seg, ahmad2022joint} of multi-person are the key tasks for human-related visual tasks. 
PersonLab~\cite{papandreou2018personlab} adopted bottom-up approach, which first extracts multiple intermediate feature maps and associates those features to get instance-wise segmentation and pose. Pose2seg~\cite{zhang2019pose2seg} used the human poses instead of the bounding boxes to normalize the interest regions for better alignment and performed the target task for each proposal. PosePlusSeg~\cite{ahmad2022joint} used a shared backbone to get intermediate maps for each task and each task pipeline combines those maps to get the results. Unlike all these convolution-based methods mentioned above, we use Transformer decoder architecture with human-centric queries to perform these tasks simultaneously.


\section{Method}
\label{sec:method}
We propose a novel human-centric query design which allows to contain various information of humans so that the human-related tasks such as detection, segmentation and pose estimation are jointly performed. It allows a light-weightened computational model with a multi-task manner, while achieving comparable accuracy in each task. The human-centric query is employed for the prediction of each task and the structural information is transferred into deformable attention module seamlessly. We first overview our unified network architecture for human-centric multi-task learning (\cref{ssec:arch_overview}), and then present our human-centric query design (\cref{ssec:query_design}) and its utilization in detail (\cref{ssec:downstream} and \cref{ssec:deformable}).

\subsection{Overall architecture}
\label{ssec:arch_overview}
The overall architecture (\cref{fig:overall}) consists of three components: image feature extractor, Transformer decoder and task-specific heads. The image feature extractor takes an image as input and produces multi-resolution features as output. The features are fed into the Transformer decoder to be attended. The Transformer decoder processes the human-centric queries to have various information of human instances, and the queries are fed into each light-weight task-specific head for final prediction.

\begin{figure}[t]
\begin{center}
\includegraphics [width=1.0\linewidth] {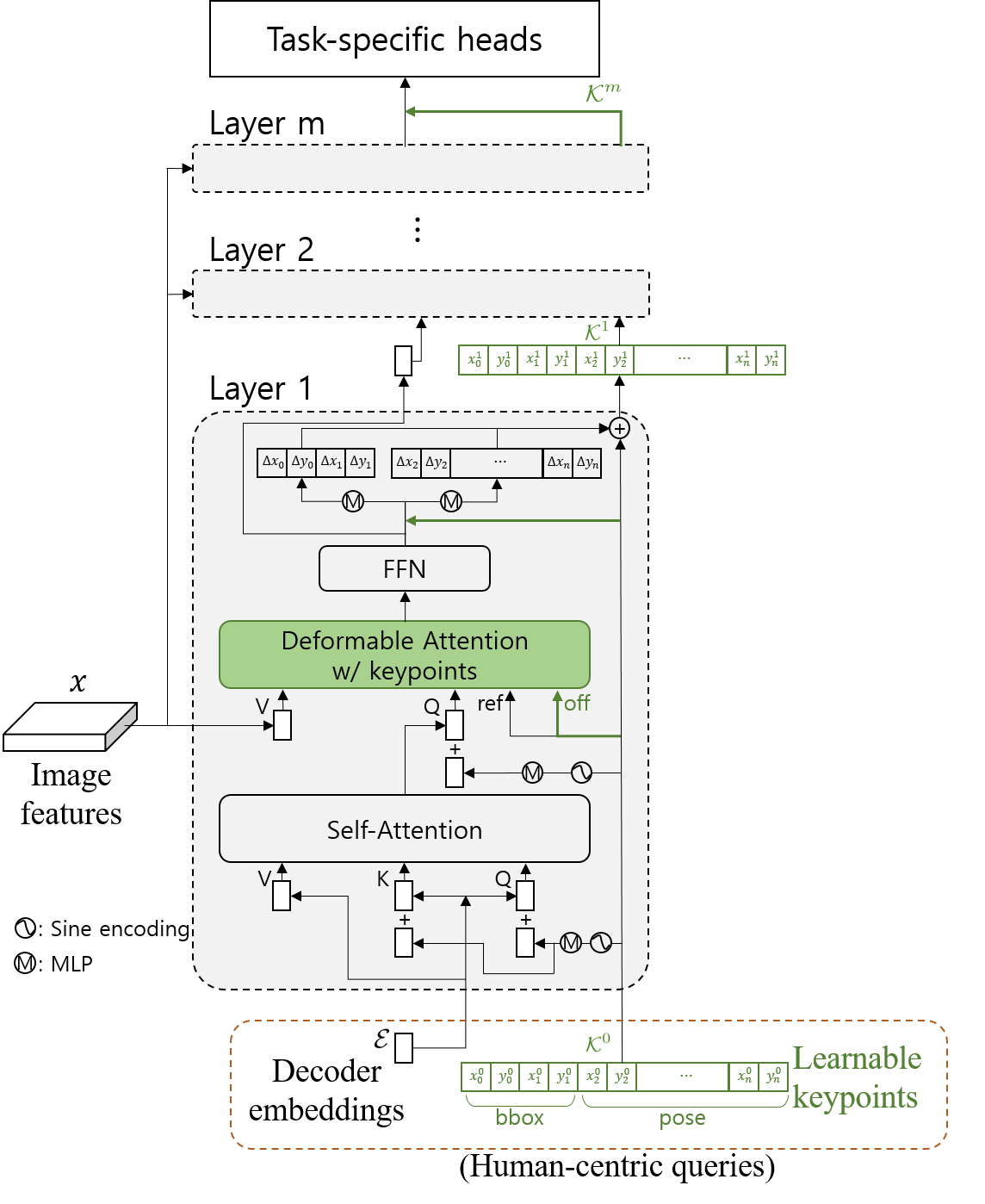}
\end{center}
\vspace{-16pt}
  \caption{Our transformer decoder architecture
}
\label{fig:decoder}
\vspace{-8pt}
\end{figure}

\subsection{Human-centric query design}
\label{ssec:query_design}
Since a single query vector for training multi-tasks represents tangled information and usually causes performance degradation, we design a novel human-centric query to decouple the structural information from the tangled embedding vector.
The proposed human-centric query consists of two distinct parts, \textit{decoder embeddings} and \textit{learnable keypoints} (\cref{fig:decoder}). 
The decoder embedding $\CONQ_q^l\in\REAL^{1 \times D}$ is the representation vector for each human instance where $q$ and $l$ are indices for query and layer, respectively, and $D$ is the hidden dimension for each query.
The learnable keypoints $\LEK_{q}^{l}\in\REAL^{1 \times 2(n+1)}$ are the decoupled positional and structural information of a human instance and defined as
\begin{equation*}
\begin{aligned}
    \LEK_q^l &=\left[ x_{q,0}^l,\,y_{q,0}^l,\,\cdots,\,x_{q,n}^l,\,y_{q,n}^l\right]\\
    &=\left[ \PTS_{q,0}^l,\,\cdots,\,\PTS_{q,n}^l\right]\,,
\end{aligned}
\end{equation*}
where $\PTS_{q,i}^l := (x_{q,i}^l,\,y_{q,i}^l)$ is a 2D coordinate of the $i\,$th keypoint.
Emphasizing that the formulation is identical for all queries and layers, we omit the indices $q$ and $l$ without loss of generalities.

The learnable keypoints $\LEK$ contains two different types of information; a \textit{bbox} part $(\PTS_0,\,\PTS_1)$ and a \textit{pose} part $(\PTS_2,\,\ldots,\,\PTS_n)$. Keypoints $\PTS_0$ and $\PTS_1$ in the \textit{bbox} part give $xy$-coordinates of left-top and right-bottom of the bounding box, respectively. These two diagonal points are sufficient to represent a given bounding box and we define the center and side lengths of the bounding box using these two points:
\begin{align}
    \PTS_c = \frac{\PTS_0 + \PTS_1}{2} \quad \mathrm{and} \quad \mathbf{d} = \PTS_1 - \PTS_0\,. \label{Def:bbox}
\end{align}
Unlike the \textit{bbox} part, keypoints in the \textit{pose} part are regarded as coordinates in a canonical space defined by the bounding box, i.e., the set $\JNT$ of $xy$-coordinates of joints in the human pose are computed as
\begin{align}
    \JNT = \left\{\PTS_0 + \PTS_i T \mid i=2,\,\ldots,\,n\right\}\,, \label{Eq:keypoints}
\end{align}
where $T:=\mathrm{diag}(\mathbf{d})$ is a dilation operator. We represent the joint cooridnates of pose in the canonical space which is normalized by the box. It reduces the pose variance according to the size of the human so that it lessens the burden of the network. The effectiveness of the canonical space can be found in the supplementary material.

Inspired by the DAB-DETR~\cite{liu2022dab}, the corresponding structural embedding $\POSQ\in\REAL^{1 \times D}$ is obtained by successive operations over $\LEK$. First, a sine encoding $\sigma\colon\REAL\rightarrow\REAL^{1 \times D'}$ maps each element of $\LEK$ to a vector and then, a multi-layer perceptron $\mathrm{MLP}$ is applied. In other words,
\begin{equation}
\begin{aligned}
    \POSQ &= \mathrm{MLP}\left( \sigma(\LEK) \right)\\
    &= \mathrm{MLP}\left(Cat\left( \sigma(x_0),\,\sigma(y_0),\,\ldots,\,\sigma(x_n),\,\sigma(y_n)\right)\right)\,,
\end{aligned}
\label{Eq:embedding}
\end{equation}
where $Cat$ is a concatenation operator along with the last dimension. Here, $\textrm{MLP}$ is a three-layer perceptron:
\small
\begin{equation*}
\begin{aligned}
    \mathrm{MLP}(\mathbf{x}) = \mathrm{ReLU}\left(\mathrm{ReLU}\left(\mathbf{x}W_1 + \mathbf{b}_1\right)W_2 + \mathbf{b}_2\right)W_3 + \mathbf{b}_3\,,
\end{aligned}
\end{equation*}
\normalsize
where $W_1\in\mathbb{R}^{2(n+1)D'\times{D}}$, $W_2,\,W_3\in\mathbb{R}^{D\times{D}}$, and $\mathbf{b}_1,\,\mathbf{b}_2,\,\mathbf{b}_3\in\mathbb{R}^{1 \times D}$ are learnable weights and biases.

One notable point is that our learnable keypoints carries salient coordinates for not only bounding box but also joints of human pose. This allows each human-centric query to become an expert on position and structure of the corresponding human object.

\subsection{Query utilization in task-specific heads}
\label{ssec:downstream}
Previous works do not use a learnable query as an input of task-specific head networks, even though the query contains high-level information pre-computed from previous layers, such as the coordinates of a bounding box (and also the joints of a pose in our approach). 
We simply feed the concatenation of the coordinates along with decoder embeddings into task-specific heads as the form of conditional information. By doing so, each task head can consider useful information in other tasks for improving its accuracy.

In \cref{fig:decoder}, regarding object detection and pose estimation tasks, as learnable keypoints themselves are the results of the tasks, their task heads perform in every layer in the transformer decoder. Other task heads perform only once, after the transformer decoder.
For all these task heads, we provide the information of learnable keypoints.

\paragraph{Pose estimation head}
While other task-specific heads have conventional structures similar to DETR~\cite{carion2020detr} and MaskFormer~\cite{cheng2021maskformer}, our pose estimation head network has a simple architecture, distinguished from previous pose estimation methods. Our head network receives object queries and produces the coordinates of pose joints directly. Off-the-shelf pose estimation methodologies~\cite{chen2018cascaded,zhang2019pose2seg,shi2022pose} used auxiliary expedients such as bounding box cropping \& resizing or heat map extraction. Although these methods has advantages in terms of model performance, it has limitations in reducing computational cost. On the other hand, thanks to the advantages inherited from object-centric queries, the proposed method uses a vector-to-vector head network which directly regresses joints of pose from the human-centric query. This makes the computational cost of the head network small compared to the conventional task head receiving a cropped image and producing a heat map (\eg $304$-to-$34$ in our head and $256\times256\times3$-to-$56\times56\times17$ in \cite{chen2018cascaded} for each instance).

\begin{figure}[t]
\begin{center}
\includegraphics [width=1.0\linewidth] {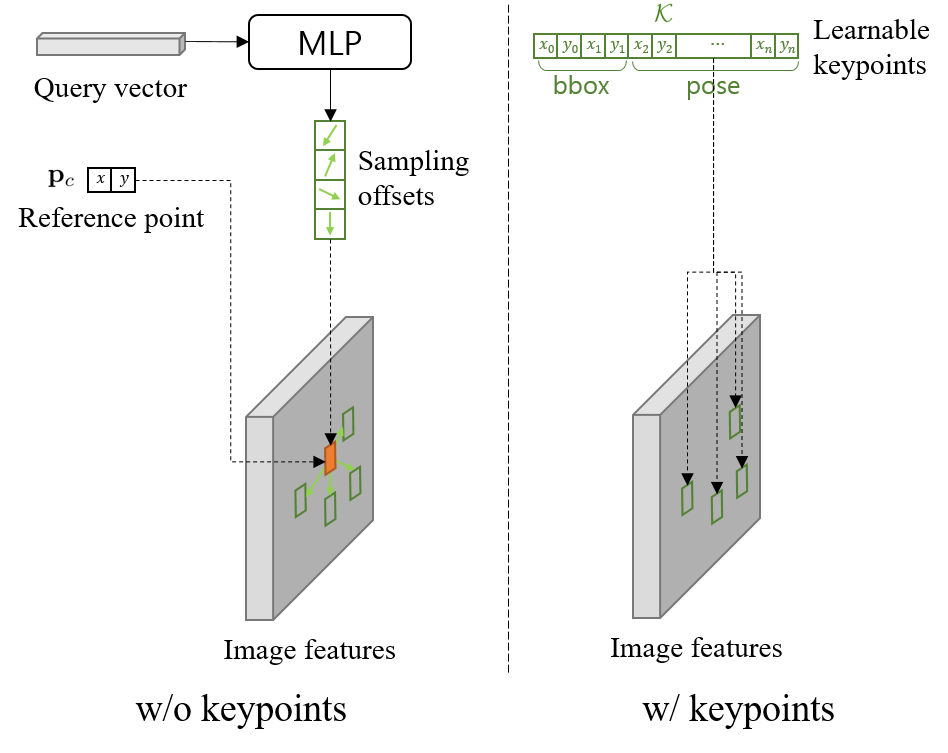}
\end{center}
\vspace{-16pt}
  \caption{Sampling locations of deformable attention w/o and w/ keypoints.
}
\label{fig:deformable}
\vspace{-8pt}
\end{figure}

\subsection{Query utilization in deformable attentions}
\label{ssec:deformable}

Our learnable query carries keypoint coordinates with high accuracy for a bounding box and a pose, and keypoints refer to salient points by definition. If we reuse such information as the sampling points in an attention module, we can reduce redundant computations for computing attention.
From this perspective, we directly use our learnable keypoints as sampling locations in a deformable attention layer.
To handle potential regions that cannot be covered by the keypoints, we also use original sampling locations together, predicted by an MLP as done in Deformable DETR~\cite{zhu2021deform}.

\paragraph{Deformable Attention with Keypoints}
Let $m$ be an attention head index and let $Q\in\REAL^{1\times D}$ be the query vector which is sum of $\POSQ$ and an output of self-attention module. The set of proposed sampling locations $\SAMLOC_{m}$ is a union of two subsets; the set $\mathcal{D}_{m}$ of Deformable DETR sampling locations and the set $\JNT_{m}$ of joint coordinates for \textit{pose}~(Fig.~\ref{fig:deformable}). The sampling locations in $\mathcal{D}_{m}$ are decided by sampling offsets $\Delta\PTS_{m}$ which are generated by MLP over the query vertor $Q$:
\begin{align}
    \mathcal{D}_{m} = \left\{\PTS_c + \Delta\PTS_{m} \mid \forall\mathrm{generated}~\Delta\PTS_{m}\right\}\,. \label{Eq:sample_deform}
\end{align}
Note that the center of the bounding box $\PTS_c$ in Eq.~\eqref{Def:bbox} plays role of the reference point. The joint coordinates for each head $\JNT_{m}$ are equally split from $\JNT$ in Eq.~\eqref{Eq:keypoints}. For the number $\nhead$ of attention heads and the number $\noff$ of all sampling locations in each head, let $x\in\REAL^{H \times W \times D}$ be an image feature map, and let $W_m\in\REAL^{D\times D/\nhead}$ be a learnable weight. The value $V_{m}\in\REAL^{\noff\times D/\nhead}$ is defined as a stack of sampled features at $x W_m$:
\begin{align}
    V_{m} = Cat\left( \left(x W_m({ \PTS })\right)^T \right)^T\,, \label{Eq:sampling}
\end{align}
where the concatenation is applied across for all ${\PTS\in\SAMLOC_{m}}$. From this, the output of the deformable attention with keypoints can be represented as
\small
\begin{align}
    \mathrm{DeformAttKey}(Q,\,V,\,\POSQ) = Cat\left(A_{1} V_{1},\,\ldots,\,A_{\nhead} V_{\nhead}\right) W\,, \label{Eq:deform_out}
\end{align}
\normalsize
where $A_{m}\in\REAL^{1\times\noff}$ is an attention coefficient obtained by linear operator over $Q$ and $W\in\REAL^{D \times D}$ is a learnable weight. 

\paragraph{Multi-Scale Attention with Keypoints}
Eq.~\eqref{Eq:deform_out} can naturally be used to the case of multi-scale attention module. Assume that there are $\nscl$ image feature maps $x_s\in\REAL^{H_s \times W_s \times D}$ for scale index $s$. By repeatedly applying the sampling process of Eq.~\eqref{Eq:sampling} for each $x_s$, sampled feature $V_{m,s}$ is obtained for each scale $s$. The multi-scale attention value $V_{m}\in\REAL^{\noff\nscl \times D/\nhead}$ is defined as
\begin{align*}
    V_{m} = Cat\left( V_{m,1}^T,\,\ldots,\,V_{m,\nscl}^T \right)^T\,.
\end{align*}
Now, the multi-scale attention output can be obtained by the same Eq.~\eqref{Eq:deform_out} if one assumes the attention coefficient having wider dimension: $A_{m}\in\REAL^{1\times\noff\nscl}$.

Specifically, regarding 32 sampling locations per an object query, we obtain 16 locations from the keypoints and the remaining 16 locations following Eq.~\eqref{Eq:sample_deform}.

\begin{table*}[t]
\centering
\small
\scalebox{1.0}{
\begin{tabular}{ c c c c c c c }
\toprule
\multirow{2}{*}{Model} & \multicolumn{3}{c}{Components} &  \multicolumn{3}{c}{Accuracy (mAP)} \\
\cmidrule(r){2-4} 
\cmidrule(r){5-7} 
& Learnable keypoints & Query util. in T.H. & Query util. in D.A. & Det. & Pose. & Seg. \\
\toprule
BaseNet-DS & & & & 56.4 & \xmark & 51.1 \\
BaseNet-DPS & & & & 53.5 & 55.7 & 49.3 \\
\cmidrule(lr){1-7} 
HCQNet-$\alpha$ & \(\checkmark\) & &  & 55.9 & 33.3 & 50.9 \\
HCQNet-$\beta$ & \(\checkmark\) & \(\checkmark\) & & 56.8 & 60.2 & 51.5 \\
HCQNet & \(\checkmark\) & \(\checkmark\) & \(\checkmark\) & 56.1 & 64.4 & 51.7 \\
\bottomrule

\end{tabular}
}
\vspace{-6pt}
\caption{Ablation study on query design. Query util. in T.H. and Query util. in D.A. mean query utilization in task-specific heads and in deformable attentions, respectively.}
\label{tbl:ablation}
\vspace{-8pt}
\end{table*}

\section{Experimental Results}
\label{sec:experiment}

\subsection{Settings}
We use MS COCO 2017 dataset~\cite{Lin2014microsoft} for network training and validation as all labels for human pose estimation, detection, and segmentation tasks are provided. 
There might be a task-specific augmentation technique to obtain the best performance for each task. However, we applied one common data augmentation technique in \cite{cheng2022mask2former} because the results of three tasks are obtained from a single image and we need to train all tasks at the same time.
We use the AdamW~\cite{loshchilov2018decoupled} optimizer with the initial learning rate of $10^{-4}$ and the batch size of 16. We train each model for $368{,}750$ iterations, and apply a learning rate decay at the two iterations of $327{,}778$ and $355{,}092$ with the decay value of $0.1$. 

\paragraph{Loss function}
We use a binary cross-entropy loss for human classification. For the segmentation, we use a binary cross-entropy and the dice loss~\cite{milletari2016vnet} as was done in \cite{cheng2022mask2former}. For the regression of the bounding box and joint positions, we use a RLE loss function~\cite{li2021rle} because RLE is known to be better than $L_1$ loss for regression problems. 
We train multiple tasks together with our unified architecture. To this end, a total training loss is defined with the weighted summation of multiple loss functions for the tasks.
Regarding bipartite matching, we use the classification and the mask loss~\cite{cheng2022mask2former} instead of using all the task losses.
We empirically found that it is sufficient for matching.
More details of the experimental setting can be found in the supplementary material.

\begin{figure}[t]
\begin{center}
\includegraphics [trim={0 0 0 40},clip,width=1.0\linewidth] {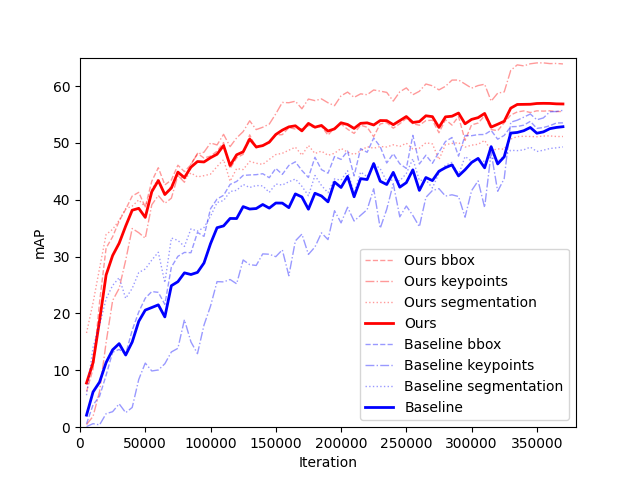}
\end{center}
\vspace{-18pt}
  \caption{Training accuracy graph (Baseline: BaseNet-DPS, Ours: HCQNet).
}
\label{fig:training_graph}
\vspace{-6pt}
\end{figure}

\begin{figure*}[t]
\centering
\small
\setlength\tabcolsep{1 pt}
  \begin{tabular}{ccccccc}
    \rotatebox[origin=l]{90}{\,  Mask2Former}  &
  	\multicolumn{1}{l}{\includegraphics[width=0.10\linewidth,height=0.13\linewidth]{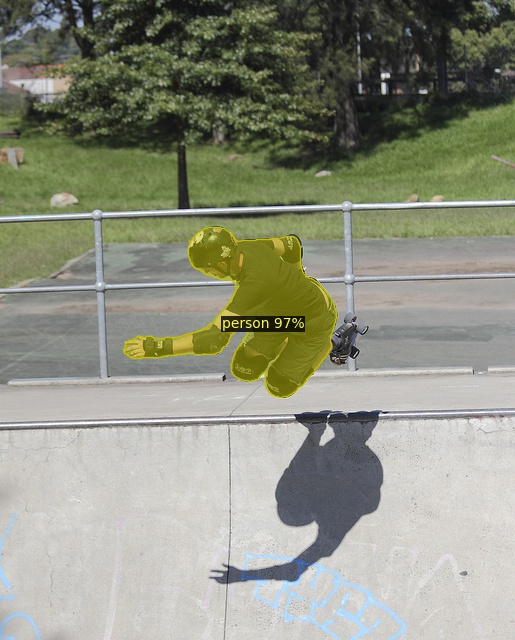}} &
  	\multicolumn{1}{l}{\includegraphics[width=0.19\linewidth,height=0.13\linewidth]{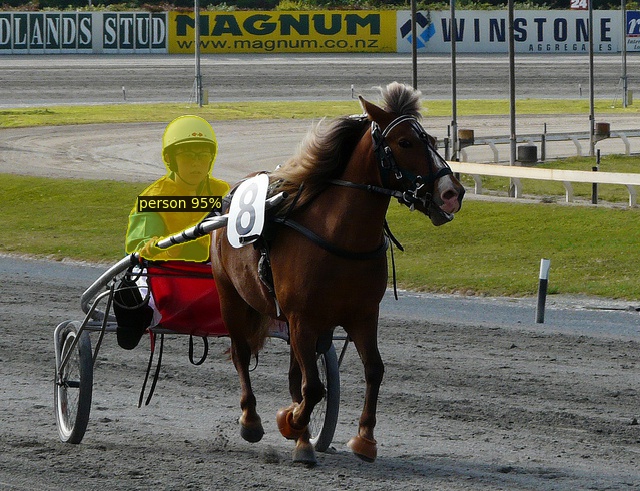}} &
  	\multicolumn{1}{l}{\includegraphics[width=0.19\linewidth,height=0.13\linewidth]{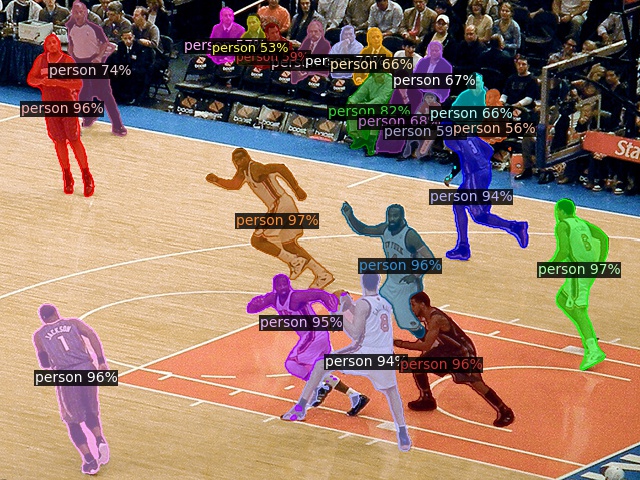}} &
  	\multicolumn{1}{l}{\includegraphics[width=0.19\linewidth,height=0.13\linewidth]{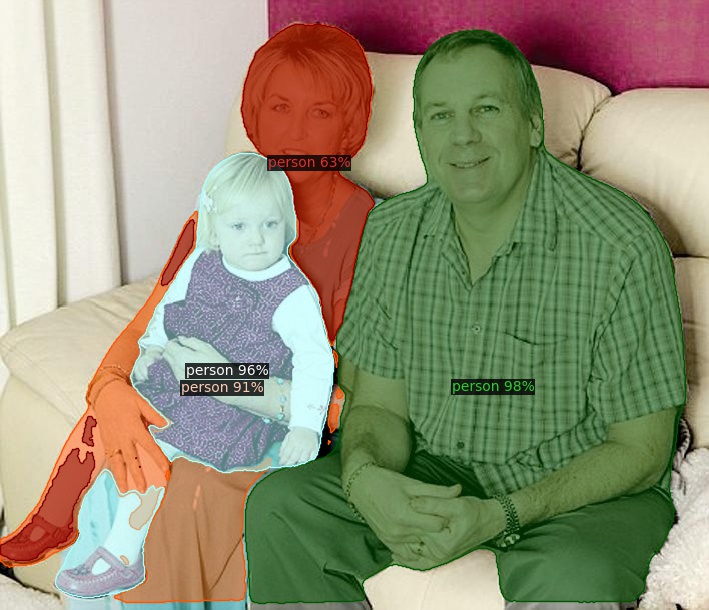}} &
  	\multicolumn{1}{l}{\includegraphics[width=0.10\linewidth,height=0.13\linewidth]{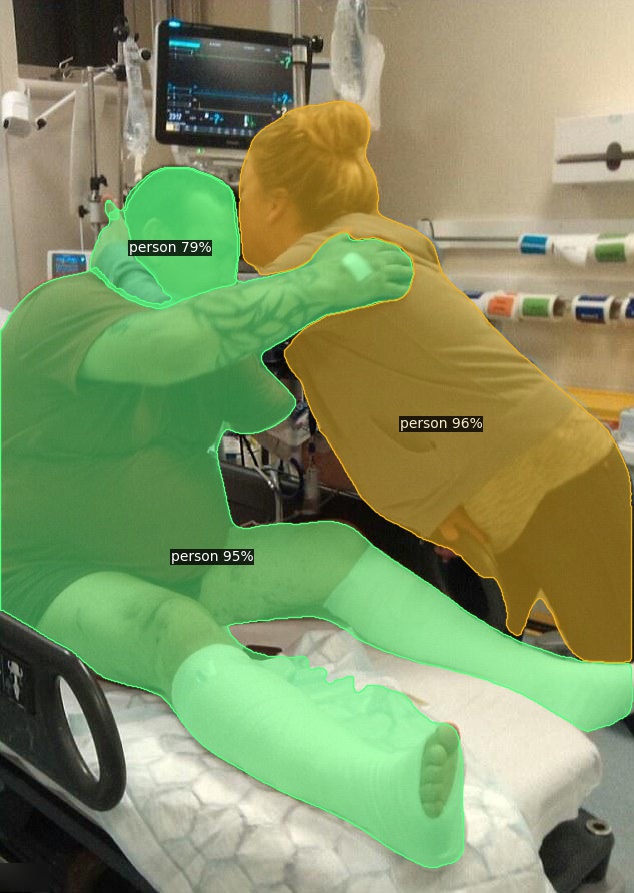}} &
  	\multicolumn{1}{l}{\includegraphics[width=0.10\linewidth,height=0.13\linewidth]{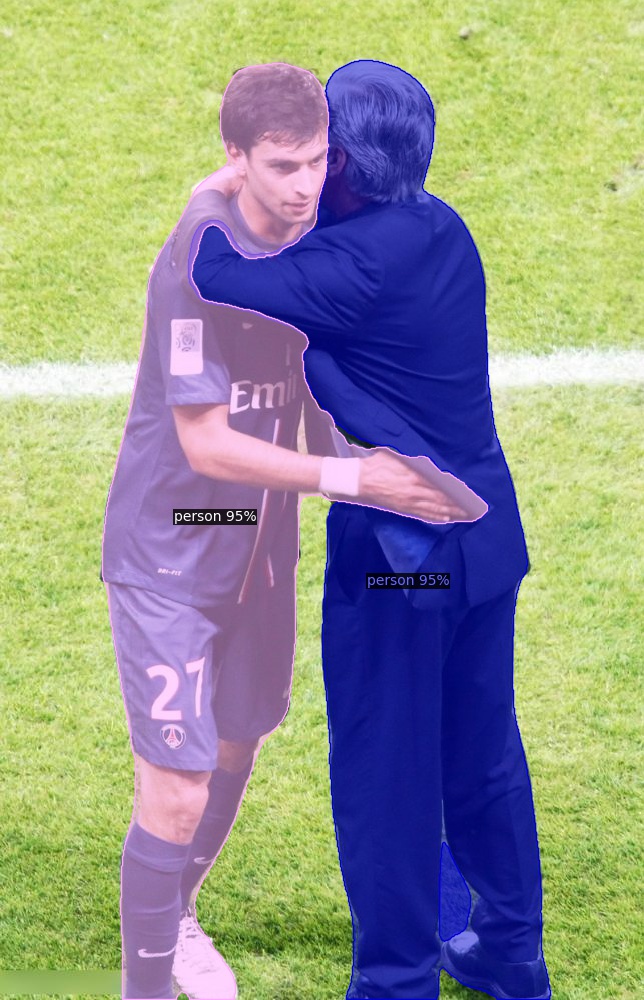}}\\

    \rotatebox[origin=l]{90}{  BaseNet-DPS} &
  	\multicolumn{1}{l}{\includegraphics[width=0.10\linewidth,height=0.13\linewidth]{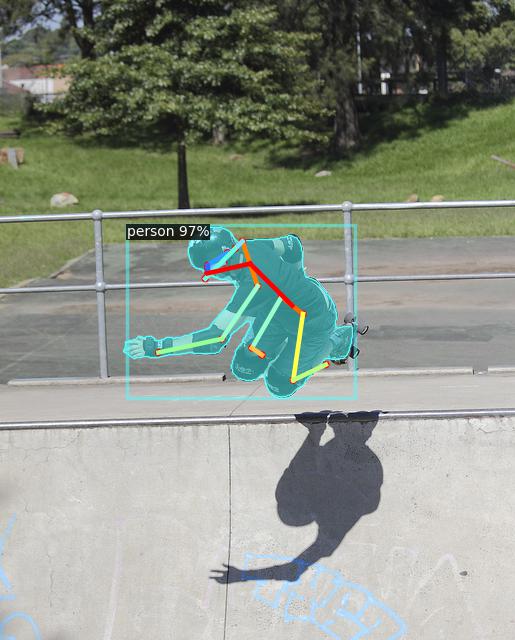}} &
  	\multicolumn{1}{l}{\includegraphics[width=0.19\linewidth,height=0.13\linewidth]{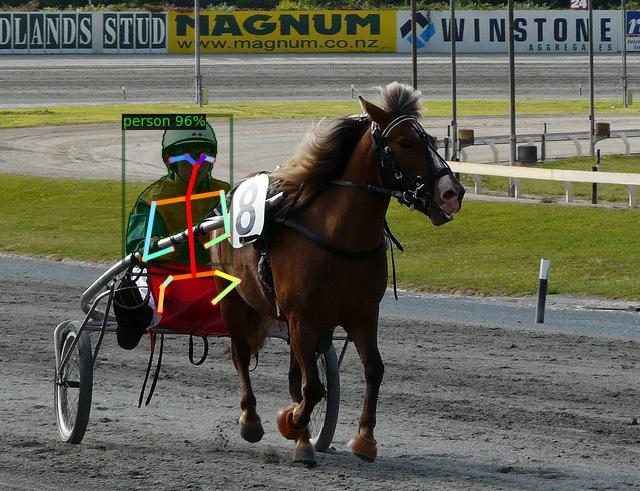}} &
  	\multicolumn{1}{l}{\includegraphics[width=0.19\linewidth,height=0.13\linewidth]{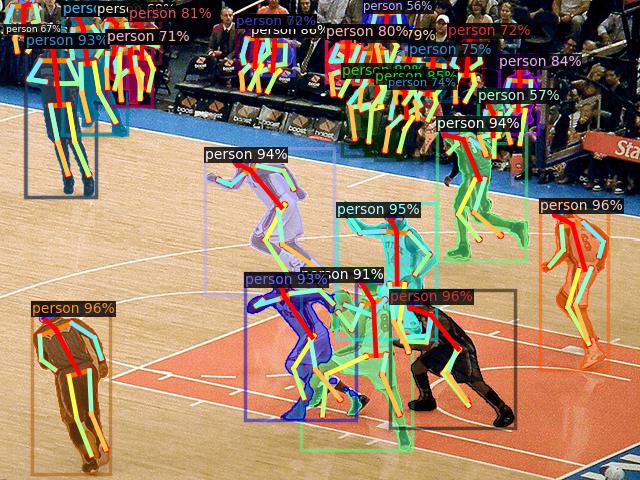}} &
  	\multicolumn{1}{l}{\includegraphics[width=0.19\linewidth,height=0.13\linewidth]{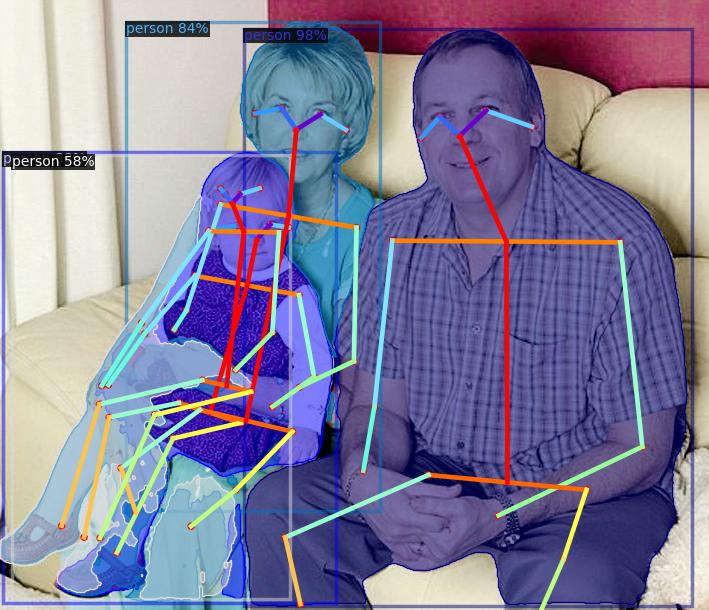}} &
  	\multicolumn{1}{l}{\includegraphics[width=0.10\linewidth,height=0.13\linewidth]{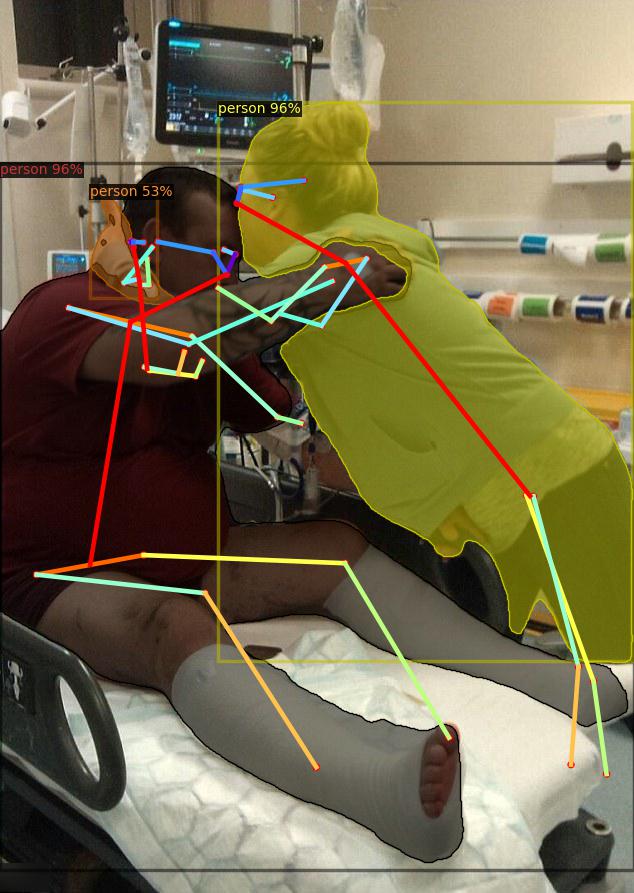}} & 
  	\multicolumn{1}{l}{\includegraphics[width=0.10\linewidth,height=0.13\linewidth]{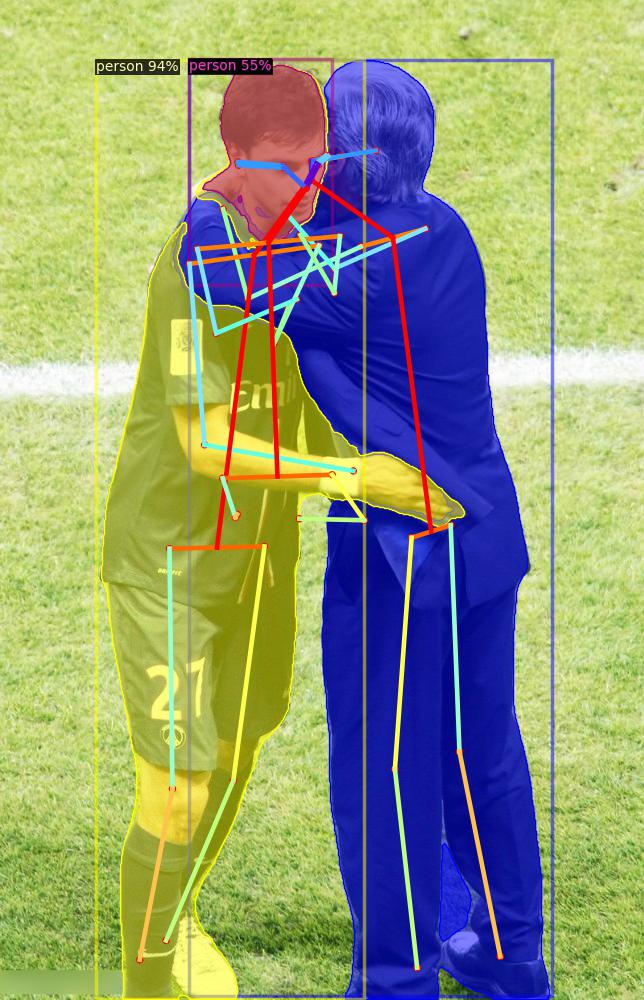}}\\

    \rotatebox[origin=l]{90}{  \, \, \, HCQNet} &
    \multicolumn{1}{l}{\includegraphics[width=0.10\linewidth,height=0.13\linewidth]{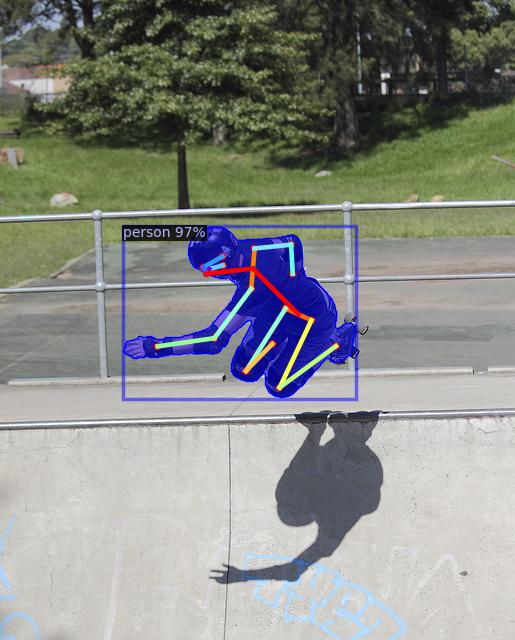}} &
  	\multicolumn{1}{l}{\includegraphics[width=0.19\linewidth,height=0.13\linewidth]{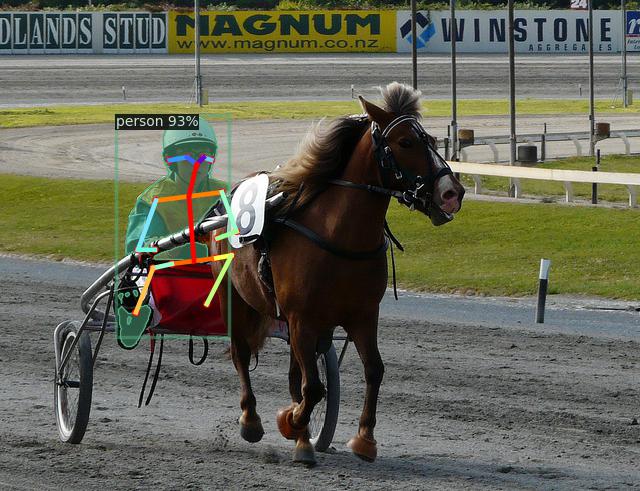}} &
	\multicolumn{1}{l}{\includegraphics[width=0.19\linewidth,height=0.13\linewidth]{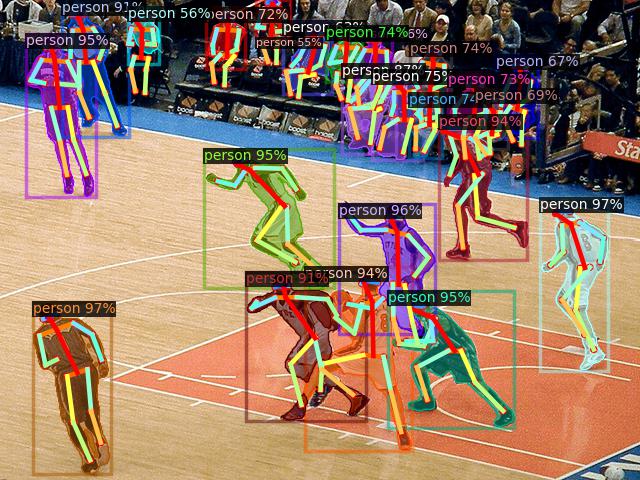}} &
	\multicolumn{1}{l}{\includegraphics[width=0.19\linewidth,height=0.13\linewidth]{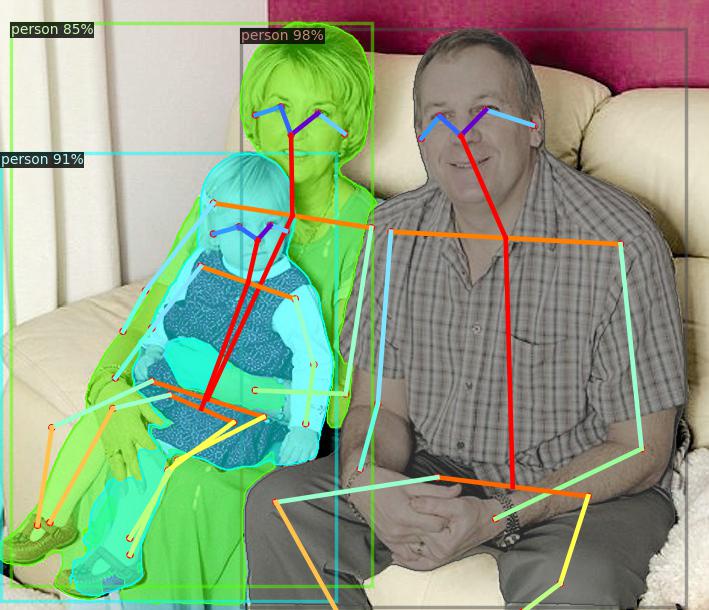}} &
  	\multicolumn{1}{l}{\includegraphics[width=0.10\linewidth,height=0.13\linewidth]{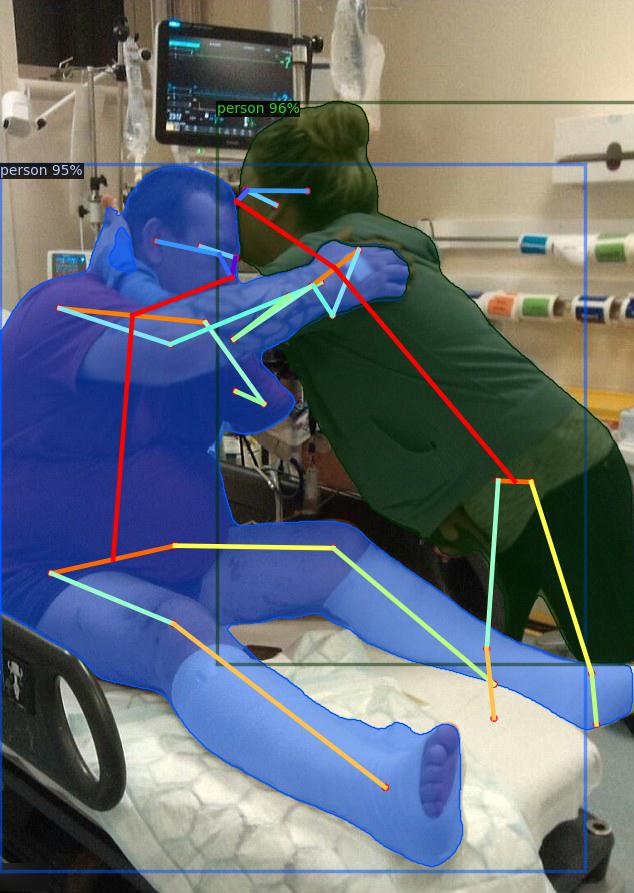}} &
  	\multicolumn{1}{l}{\includegraphics[width=0.10\linewidth,height=0.13\linewidth]{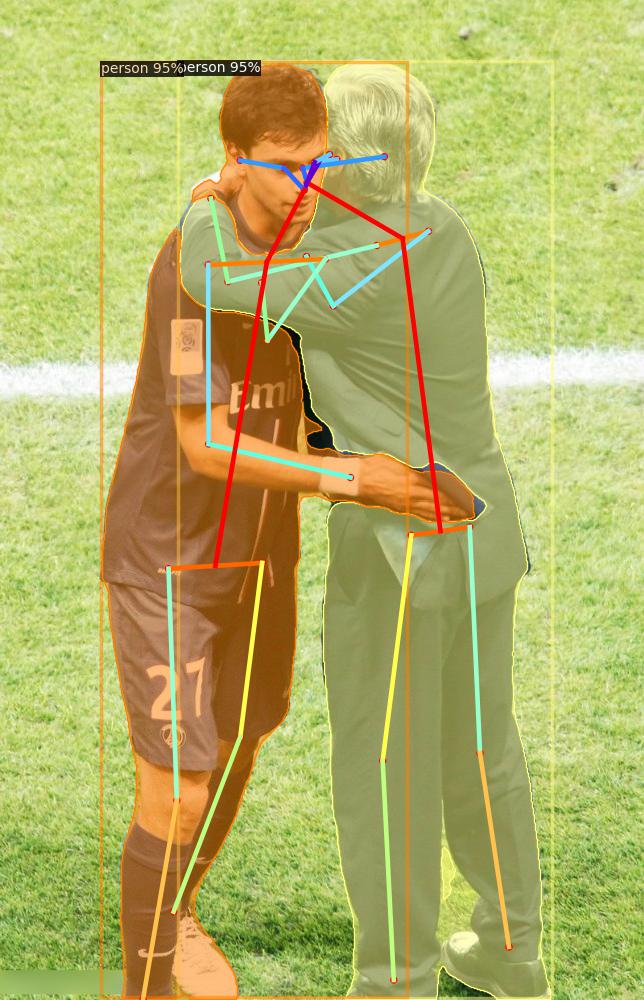}}\\

  \end{tabular}

\vspace{-12pt}
  \caption{Qualitative results. Left three columns are from the COCO 2017 Person dataset~\cite{Lin2014microsoft}, others are from the OCHuman dataset~\cite{zhang2019pose2seg}.
}
\label{fig:qualitative}
\vspace{-6pt}
\end{figure*}

\subsection{Component analysis}
\label{sec:component_anal}
\paragraph{Ablation study}
Our baseline network architecture is based on Mask2Former~\cite{cheng2022mask2former} for segmentation using the Swin-B~\cite{liu2021swin} backbone, and we replace their cross-attention layers with deformable attention layers. To predict reference points for deformable attention, we replace learnable query with the \textit{bbox} part of learnable keypoints.
We define two models BaseNet-DS and BaseNet-DPS as baseline (\Cref{tbl:ablation}). The BaseNet-DS conducts only detection and segmentation tasks with above settings. 
In addition to BaseNet-DS, we consider na\"ive multi-tasking with an additional human pose estimation task. We add a head network and a RLE loss fucntion for human pose estimation to the baseline model, and we consider this model as a baseline multi-tasking model~(BaseNet-DPS).

In this ablation study~(\Cref{tbl:ablation}), from the baseline models, we add our components of 1) learnable keypoints including also the joint coordinates of a pose, 2) query utilization in task-specific heads (Query util. in T.H.), and 3) query utilization in deformable attentions (Query util. in D.A.) one by one, and analyze their performance.

\begin{table}[t]
\centering
\small
\scalebox{1.0}{
\begin{tabular}{c c c c }
\toprule
 \multirow{2}{*}{Variants of Learnable keypoints}  &  \multicolumn{3}{c}{Accuracy (mAP)} \\
\cmidrule(r){2-4} 
 & Det. & Pose. & Seg. \\
\toprule
Canonical space coordinate & 56.1 & 64.4 & 51.7\\
Image space coordinate& 56.2 & 63.0 & 51.5\\
Canonical space embedding  & 56.2 & 63.7 & 51.8\\
\bottomrule
\end{tabular}
}
\vspace{-4pt}
\caption{Effect of different forms of leranable keypoints used in task-specific heads (segmentation and detection).}
\label{tbl:query_form}
\vspace{-4pt}
\end{table}

Simply adding a task causes the degradation of overall performance showing a difficulty of learning shared representation for multi-tasking~(BaseNet-DS vs. BaseNet-DPS).
When we employ our query design, the accuracy of detection and segmentation increase (BaseNet-DPS vs. HCQNet-$\alpha$). It shows that explicit separation of mixed information can improve accuracies.
However, in this case, the accuracy of human pose estimation drops significantly, because decoder embeddings now do not contain pose information sufficiently.

When we provide learnable keypoints to task-specific heads, the accuracies of all tasks increase because the positional and structural information in the learnable keypoints can complement each other of the tasks (HCQNet-$\alpha$ vs. HCQNet-$\beta$).
We found that this component is more effective in pose estimation than detection even though they have a similar prediction process using regression and iterative refinement.
We guess that, in detection, a bounding box has simpler representation so that the prediction head can estimate a proper displacement without the bounding box predicted in the previous layer, but not \textit{vice versa} in pose estimation.

By using the learnable keypoints in deformable attentions, we can expect additional performance gain (HCQNet-$\beta$ vs. HCQNet).
In our experiment, this component significantly improves the accuracy of human pose estimation, 
but the gain in segmentation is smaller and it slightly decreases the accuracy of detection.
Overall, our all components (HCQNet) bring significant performance improvement for all target tasks, compared to the baseline multi-tasking model (BaseNet-DPS).

\paragraph{Training speed}
In addition to improvement of overall accuracy in all tasks, we found that the training speed of our approach is much faster than the baseline multi-tasking model (\cref{fig:training_graph}). Specifically, HCQNet trained for 100k iterations shows higher average accuracy than BaseNet-DPS trained for 300k iterations, showing more than three times faster training speed. It implies that our approach can be useful for training a larger multi-task model.

\paragraph{Qualitative comparison}
The effects of our proposed approach can be also found in qualitative results (\cref{fig:qualitative}).
As shown in the second column of \cref{fig:qualitative}, while task-specific method (Mask2Former) and na\"ive multi-tasking model (BaseNet-DPS) suffer from segmenting an occluded object, our model (HCQNet) can produce better segmentation result by jointly utilizing the information of other tasks encoded in our HCQ.

\paragraph{Variants of the learnable keypoints}
In the previous experiment, our proposed components successfully improve learnable keypoints and the performance of various tasks.
Meanwhile, different forms of learnable keypoints as conditional information may lead more performance improvement because there can be a suitable form of the conditional information for each task.
In this experiment, we explore the possibility.
We empirically found that the current form using the coordinates of pose joints in the canonical space is suitable for learning pose estimation. 
Therefore, we test different forms for only object detection and segmentation tasks.
For the tasks, we test other forms such as the joint coordinates in the image space or keypoint embedding instead of coordinate. For keypoint embedding, we use the same process of obtaining structural embedding used in (\cref{Eq:embedding}).

The canonical space coordinate has better accuracy in pose estimation compared to the other variants while having the comparable accuracy on detection and segmentation tasks (\Cref{tbl:query_form}). This implies that our coordinate information in learnable keypoints can be generally applied to various tasks.


\subsection{Comparison}


To show the effectiveness of our approach, we compare our method with various task-specific models~\cite{carion2020detr,cheng2021maskformer,cheng2022mask2former,liu2022dab,shi2022pose} and multi-task models~\cite{he2017mask,zhang2019pose2seg,ahmad2022joint} handling a part of object dectection, instance segmentation, and human pose estimation tasks.
We use a residual network~\cite{he2016residual} (R-50 and R-152), a feature pyramid network~\cite{lin2017feature} (fpn), and Swin-Transformer~\cite{liu2021swin} (Swin-B) as backbone models, pretrained on the ImageNet-1K dataset~\cite{deng2009imagenet}. 
As there is no approach to handle three tasks together, we divide the tasks into two groups in terms of evaluation protocol for fair comparison. Specifically, when we evaluate tasks including pose estimation, the instances with a small size ($< 32\times32$ pixels) are excluded in computing mAP, as was done in \cite{zhang2019pose2seg,ahmad2022joint,shi2022pose}.
Note that once we trained our HCQNet jointly for all tasks, we use it to evaluate with two types of protocals.

\begin{table}[t]
\centering
\small
\scalebox{1.0}{
\begin{tabular}{ c c c c c }
\toprule
\multirow{2}{*}{Model} & \multirow{2}{*}{Backbone} & \multicolumn{3}{c}{Accuracy (mAP)} \\
\cmidrule(r){3-5} 
& &  Det. & Pose. & Seg.\\
\toprule
Mask R-CNN$^*$~\cite{he2017mask} & R-50 & 52.0 & \xmark & 43.6\\
DETR$^*$~\cite{carion2020detr} & R-50 & 52.8 & \xmark & \xmark \\
DAB-DETR$^*$~\cite{liu2022dab} & R-50 &51.8 &\xmark &\xmark\\
HCQNet & R-50 & 52.6 & 60.4 & 49.1 \\ 
\cmidrule(lr){1-5} 
Mask2Former~\cite{cheng2022mask2former} & Swin-B & \xmark & \xmark & 52.0 \\ 
DAB-DETR~\cite{liu2022dab} & Swin-B & 56.4 & \xmark & \xmark \\
HCQNet & Swin-B & 56.1 & 64.4 & 51.7 \\
\bottomrule

\end{tabular}
}
\vspace{-6pt}
\caption{Comparison with state-of-the-art task specific models on the COCO 2017 Person \textit{minval} set. The asterisk~* denotes models trained for handling general classes, downloaded from the Detectron2~\cite{wu2019detectron2} and the authors' websites.  }
\label{tbl:comparison}
\vspace{-4pt}
\end{table}

\begin{table}[t]
\centering
\small
\scalebox{1.0}{
\begin{tabular}{ c c c c c }
\toprule
\multirow{2}{*}{Model} & \multirow{2}{*}{Backbone} & \multicolumn{3}{c}{Accuracy (mAP)} \\
\cmidrule(r){3-5}
& &  Det. & Pose. & Seg. \\
\toprule
Pose2Seg~\cite{zhang2019pose2seg} & R-50-fpn & \xmark & 59.9$^*$ & 55.5 \\
Pose2Seg(GT kpt) & R-50-fpn & \xmark & GT & 58.2 \\ 
PosePlusSeg~\cite{ahmad2022joint} & R-152 & \xmark & 74.4 & 56.3  \\
PETR~\cite{shi2022pose} & R-50 & \xmark & 67.4 & \xmark \\
HCQNet & Swin-B & 68.9 & 65.2 & 65.1 \\
HCQNet$_{ft}$ & Swin-B & 69.2 & 65.6 & 65.5 \\
\bottomrule

\end{tabular}
}
\vspace{-8pt}
\caption{Comparison with pose estimation methods on the COCO 2017 Person \textit{minval} set (without small person instances). 
*~Pose2Seg uses a stand-alone model for pose estimation~\cite{newell2017associative}.
}
\label{tbl:COCOPersonVal_W/Osmall}
\vspace{-4pt}
\end{table}

In \Cref{tbl:comparison}, we compare our HCQNet with detection and segmentation models~\cite{he2017mask,carion2020detr,liu2022dab,cheng2022mask2former}.
For a fair comparison, we retrain state-of-the-art task specific models with the Swin-B backbone for a human class only; DAB-DETR~\cite{liu2022dab} in object detection and Mask2Former~\cite{cheng2022mask2former} in instance segmentation. 
\Cref{tbl:comparison} shows that, even though our model runs three tasks simultaneously, it achieves comparable performance to task-specific state-of-the-art models.


In \Cref{tbl:COCOPersonVal_W/Osmall}, we compare models handling tasks including pose estimation~\cite{zhang2019pose2seg,shi2022pose,ahmad2022joint}.
In this experiment, we exclude small-sized human instances in the validation set as mentioned above.
Our model shows a significantly higher segmentation accuracy than them (\Cref{tbl:COCOPersonVal_W/Osmall}).
The pose estimation accuracy is lower than previous methods, because our models consider small objects in data augmentation, while poes estimation methods do not.
When we finetune our model on the training set focusing on larger instances, our overall performance further improves.
As our model uses the simpler architecture of our pose-specific head compared to other pose-specific approaches and is not specialized for a human pose estimation task, the accuracy of our model in pose estimation is still inferior to those of state-of-the-art pose estimation methods~\cite{shi2022pose,ahmad2022joint}.

Our model is a simple combination of multiple tasks, and not currently optimized in terms of loss balancing, inter-task-correlated loss, and other task-specific designs.
Nonetheless, our model shows comparable performance to task-specific models.
We believe that exploring them would lead to the performance improvement of our approach.
We refer the reader to our supplementary material for results on the OCHuman dataset~\cite{zhang2019pose2seg}.

\subsection{Computational cost analysis}
In previous experiments, we show that our query design enables a unified network to perform multiple task harmoniously in terms of accuracy.
In this section, we validate that the cost effectiveness of our multi-task network.
The computational costs of comparing methods are computed on an input image of $1024\times1024$.
In our network, multiple tasks share most computations in backbone, pixel decoder (corresponding transformer encoder), and transformer decoder (\Cref{tbl:computation}).
The overhead of each task head is negligible, showing the scalability of our approach in terms of increasing the number of target tasks.
In addition, it is noteworthy that performing our model alone is more cost-efficient than performing task-specific state-of-the-art models separately.

\begin{table}[t]
\centering
\small
\scalebox{0.93}{
\begin{tabular}{ c c c c}
\toprule
Model & Module &  Cost (BFlops) & Proportion (\%) \\
\toprule
\multirow{8}{*}{HCQNet}&Backbone &  363 & 69.96 \\
&Pixel decoder & 143 & 27.56 \\
&Trans. decoder & 12.60 & 2.43\\
&Class & 0.0000512 & 0.00 \\
&Mask & 0.020634 & 0.00 \\
&Box & 0.128 & 0.02 \\
&Pose & 0.135 & 0.03 \\
\cmidrule(lr){2-4} 
&Total & 521 & \underline{100} \\
\cmidrule(lr){1-4} 
Separate & DAB-DETR~\cite{liu2022dab} & 361 & 69\\
task- & Mask2Former~\cite{cheng2022mask2former} & 535 & 102\\
specific & PETR~\cite{shi2022pose} & 747 & 143\\
\cmidrule(lr){2-4} 
models & Total & 1643 & 315\\
\bottomrule

\end{tabular}
}
\vspace{-4pt}
\caption{Computational cost analysis. Backbone of all the models is Swin-B~\cite{liu2021swin}.}
\label{tbl:computation}
\vspace{-4pt}
\end{table}





\section{Conclusion}
\label{sec:conclusion}
We present a novel human-centric representation for multi-task learning. To this end, a novel query is designed that carries positional coordinates of key points and represents structural information of human instances effectively. In order to exploit the queries with pre-computed, high-level information in task-specific heads, we use the learnable keypoints as conditional input for the task heads and combine the learnable keypoints with deformable attention. As the results, our proposed model shows comparable performance to individual state-of-the art models for multiple human recognition tasks such as pose estimation, segmentation, and detection, while it consumes a significantly small computational cost.
We refer the reader to our supplementary material for discussion about the limitation and potential negative social impact of our approach.

%


{\small
\bibliographystyle{ieee_fullname}
\bibliography{egbib}
}

\clearpage
\appendix




\section{Details About Experimental Setting}
\subsection{Network architecture}
As described in Sec. 3.1 in the main paper, our network consists of image feature extractor, transformer decoder, and task-specific modules.
Our main difference in the architecture is to concentrate on the transformer decoder.
The image feature extractor is based on the Mask2Former~\cite{cheng2022mask2former}, which consists of a backbone network and a pixel decoder. We can select various backbone models such as residual network~\cite{he2016residual} and Swin-transformer~\cite{liu2021swin}. Following \cite{cheng2022mask2former}, we use multi-scale deformable attention Transformer (MSDeformAttn)~\cite{zhu2021deform} as the pixel decoder.
Regarding the transformer decoder, each individual human in the given image corresponds to each of proposed human-centric queries, and we use $100$ queries for all experiments. The decoder consists of $8$ decoder layers.
Distinct form \cite{cheng2022mask2former}, in each decoder layer, we use a deformable attention layer~\cite{zhu2021deform} as a cross-attention layer, and a self-attention layer is placed before the deformable attention layer.
Each deformable attention layer receives $3$ scales of image feature ($1/32$, $1/16$, $1/8$ of the feature resolution) from the pixel decoder.

\subsection{Loss function}
As described in Sec 4.1 in the main paper, we use four types of loss functions according to the target tasks: classification, segmentation, bbox, and pose.
We denote these loss functions as $L_{c}$, $L_{s}$, $L_{b}$, and $L_{p}$, respectively.  
Then, the total training loss is defined as
\begin{align}
L_{total} = \lambda_{c} L_{c} + \lambda_{s} L_{s} +\lambda_{b} L_{b} + \lambda_{p} L_{p}.
\end{align}
where $\lambda_{c}$, $\lambda_{s}$, $\lambda_{b}$ and $\lambda_{p}$ are the mixing weights and are set to 2, 5, 0.2, and 0.2, respectively, in our experiment.
Because there are many possible combinations of mixing weights for multiple tasks, another best combination would exist.
Still, our models with these weights show practically reasonable performance in the target tasks without sophisticated tuning to search the best hyper-parameters.

The training loss is applied to the matched pairs of instances between the prediction and the ground-truth.
We also use an auxiliary training loss by attaching the prediction layer to each transformer decoder layer, similar to \cite{cheng2022mask2former}.
The auxiliary loss is same with $L_{total}$.

\subsection{Data augmentation}
As described in Sec .4.1 in the main paper, we follow a data augmentation scheme used in \cite{cheng2022mask2former}.
Specifically, for each image, we apply random scaling to the image with the range [0.1, 2.0] and crop the scaled image with the fixed size of $1024 \times 1024$. If the scaled image is smaller than the cropping size, we apply zero-padding to right- and bottom-side of the image to produce the result image of $1024 \times 1024$.

\begin{figure*}[t]
\centering
\setlength\tabcolsep{1 pt}
  \begin{tabular}{cccccc}
  	\multicolumn{1}{l}{\includegraphics[height=0.116\linewidth]{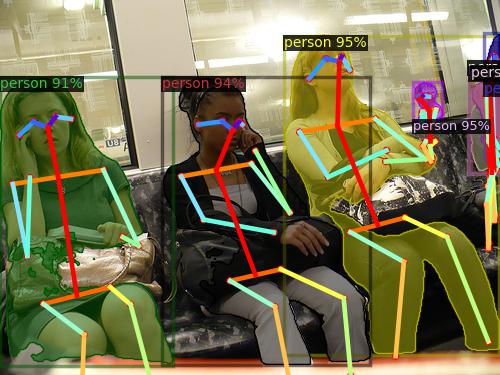}} &
  	\multicolumn{1}{l}{\includegraphics[height=0.116\linewidth]{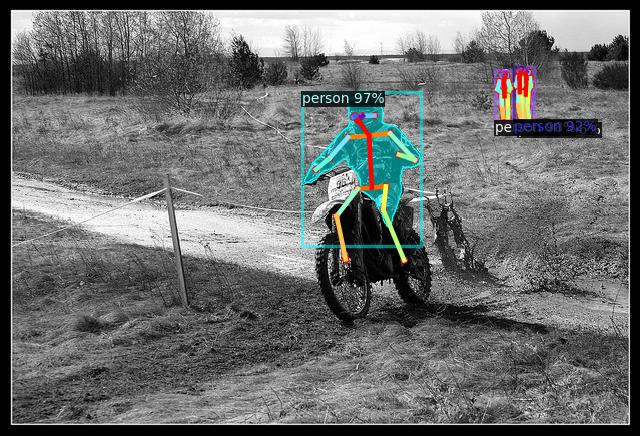}} &
  	\multicolumn{1}{l}{\includegraphics[height=0.116\linewidth]{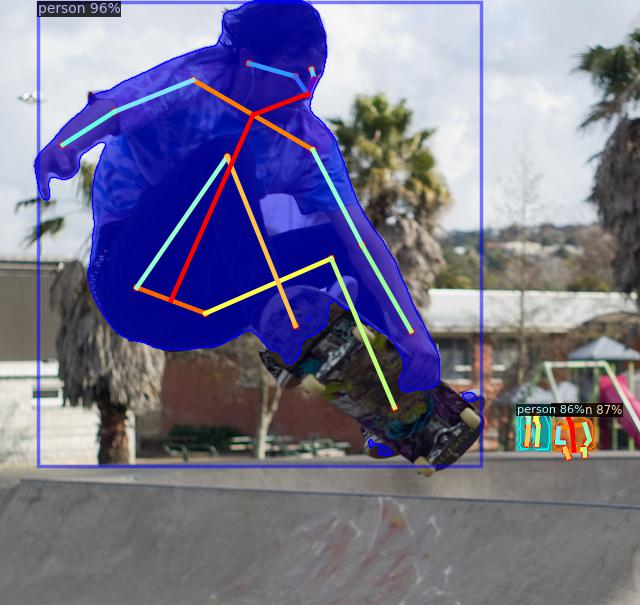}} &
  	\multicolumn{1}{l}{\includegraphics[height=0.116\linewidth]{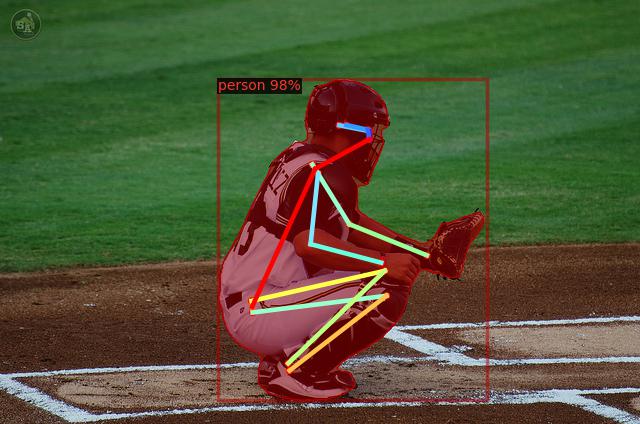}} &
  	\multicolumn{1}{l}{\includegraphics[height=0.116\linewidth]{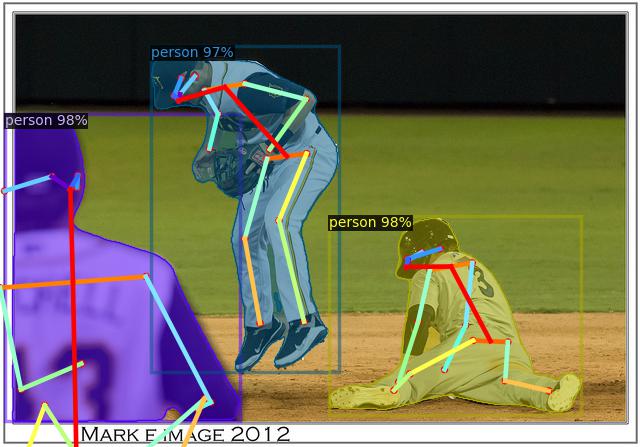}} &
  	\multicolumn{1}{l}{\includegraphics[height=0.116\linewidth]{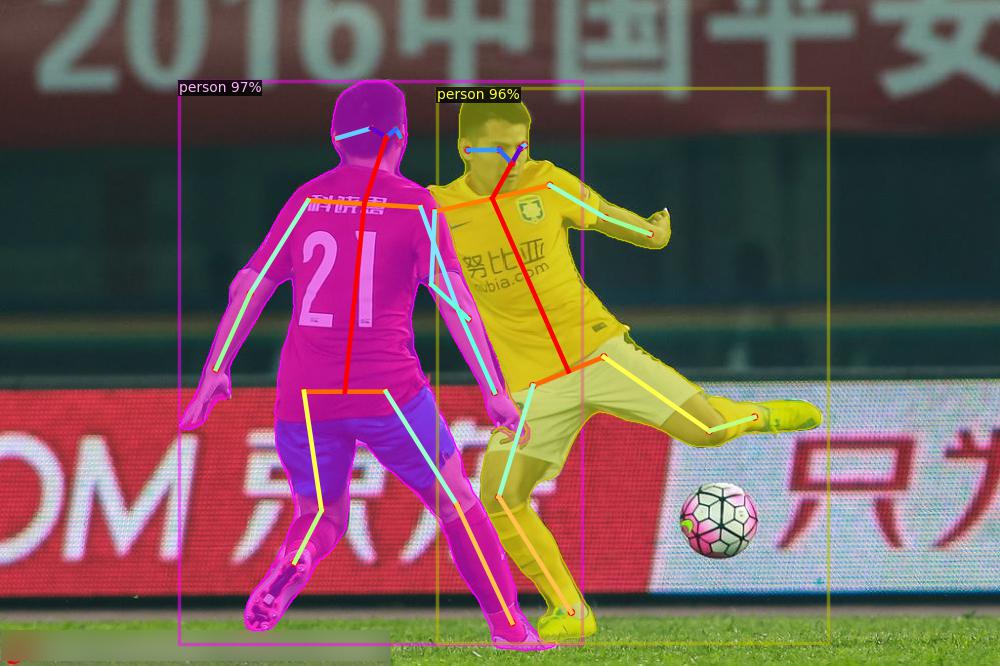}} \\
  \end{tabular}
\vspace{-12pt}
  \caption{Additional visual results of our HCQNet on scenes containing dynamic human poses. 
}
\label{fig:limitation}
\vspace{-6pt}
\end{figure*}

\begin{figure*}[t]
\centering
\setlength\tabcolsep{1 pt}
  \begin{tabular}{ccccc}
  	\multicolumn{1}{l}{\includegraphics[height=0.2\linewidth]{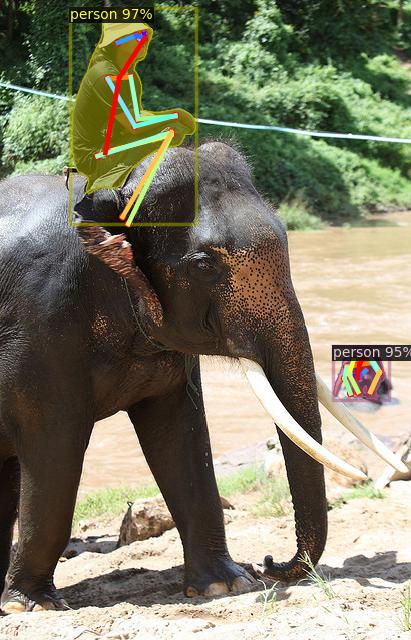}} &
  	\multicolumn{1}{l}{\includegraphics[height=0.2\linewidth]{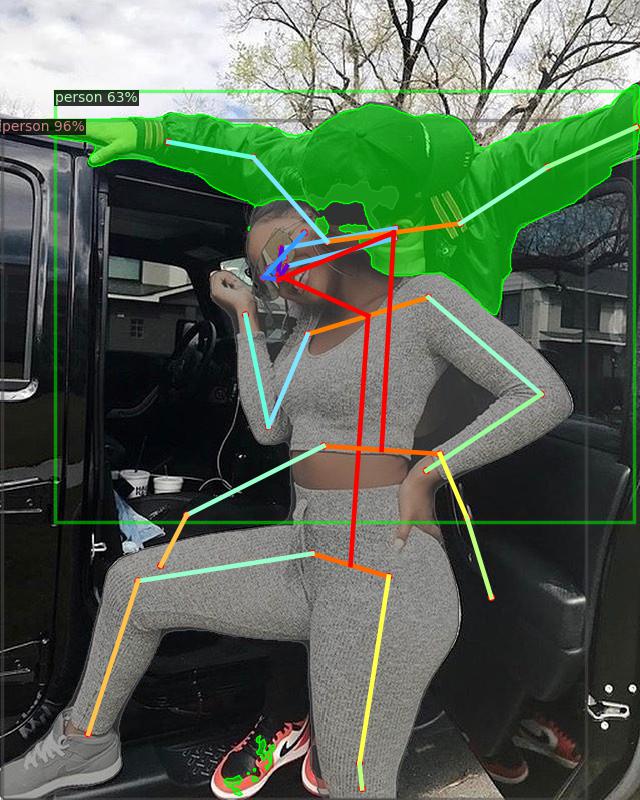}} &
  	\multicolumn{1}{l}{\includegraphics[height=0.2\linewidth]{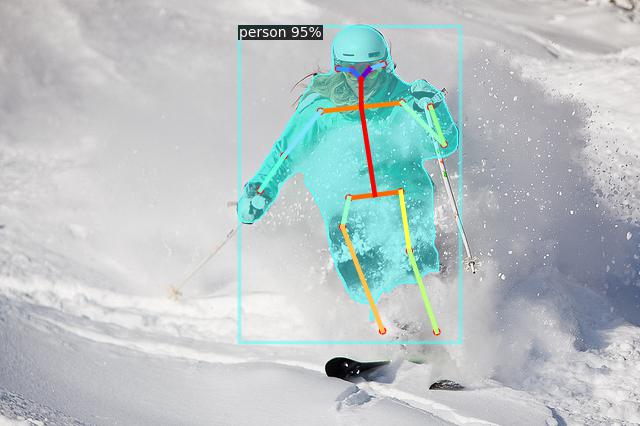}} &
  	\multicolumn{1}{l}{\includegraphics[height=0.2\linewidth]{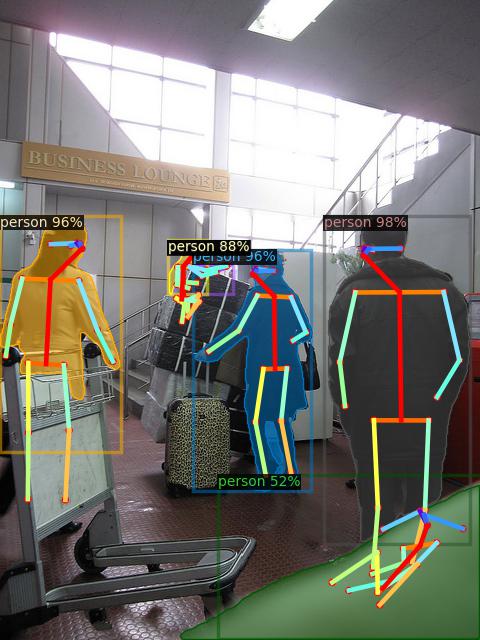}} &
  	\multicolumn{1}{l}{\includegraphics[height=0.2\linewidth]{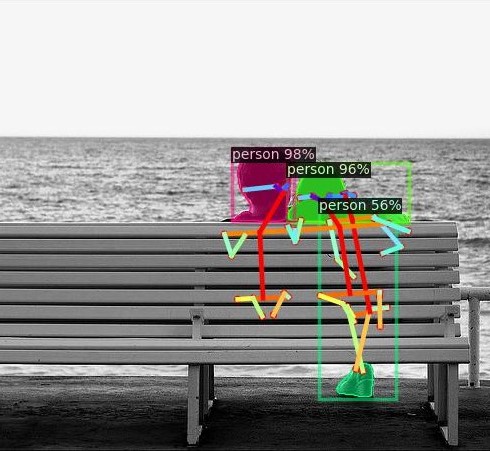}} \\
  \end{tabular}
\vspace{-12pt}
  \caption{Failure cases. 
}
\label{fig:limitation}
\vspace{-8pt}
\end{figure*}

\section{Additional Experiments}

\subsection{Canonical space of the pose part of learnable keypoints}

As described in Sec. 3.2 in the main paper, 
we normalized the coordinates of the \textit{pose part} in our learnable keypoints by the box coordinate of the \textit{bbox part}; we refer it as the canonical space.
We can also define the coordinates of \textit{pose part} in the image space instead of canonical space.
We empirically found that representing the \textit{pose part} in the image space causes a large performance drop in pose estimation and it also degrades the accuracy of two other tasks (\Cref{tbl:canonical}).

\subsection{OCHuman dataset}
In Fig. 6 in the main paper, we present several qualitative results on the OCHuman dataset~\cite{zhang2019pose2seg}.
In this section, we present quantitative results on the dataset.
Compared to Pose2Seg~\cite{zhang2019pose2seg}, our method achieves better performance in both detection and segmentation tasks (\Cref{tbl:OCHumanComparison}).
Compared to a state-of-the-art segmentation method (Mask2Former~\cite{cheng2022mask2former}), our method shows a comparable accuracy in segmentation.
Compared to our baseline multi-tasking model (BaseNet-DPS), our approach still induces significant improvements in pose estimation and segmentation on this different dataset.

We show additional visual results of our methods on the COCO and OCHuman datasets.


\begin{table}[t]
\centering
\small
\scalebox{0.9}{
\begin{tabular}{ c c c c }
\toprule
\multirow{2}{*}{Pose joint coordinates of learnable keypoints} &  \multicolumn{3}{c}{Accuracy (mAP)}\\

\cmidrule(r){2-4} 
 & Det. & Pose. & Seg. \\
\toprule
 In the image space  & 54.8 & 55.3 & 49.6 \\
 In the canonical space & 56.1 & 64.4 & 51.7 \\

\bottomrule

\end{tabular}
}
\vspace{-7pt}
\caption{Effectiveness of employing the canonical space for the pose part of our learnable keypoints. }
\label{tbl:canonical}
\vspace{-5pt}
\end{table}

\begin{table}[t]
\centering
\small
\scalebox{0.8}{
\begin{tabular}{ c c c c c c c c }
\toprule
\multirow{2}{*}{Model} & \multirow{2}{*}{Backbone} & \multicolumn{3}{c}{OCHuman Val (mAP)} &  \multicolumn{3}{c}{ OCHuman Test (mAP)} \\
\cmidrule(r){3-5} 
\cmidrule(r){6-8}
& &  Det. & Pose. & Seg. &  Det. & Pose. & Seg. \\
\toprule
Pose2Seg & R-50 & \xmark & 28.5 & 22.2 & \xmark & 30.3 & 23.8 \\ 
Mask2Former & Swin-B & \xmark & \xmark & 27.5 & \xmark & \xmark & 27.8 \\ 
BaseNet-DPS & Swin-B & 19.8 & 30.2 & 25.6 & 19.4 & 29.7 & 25.5 \\
HCQNet & Swin-B & 19.7 & 31.0 & 27.1 & 19.4 & 30.9 & 27.3 \\
\bottomrule

\end{tabular}
}
\vspace{-7pt}
\caption{Comparison on the OCHuman Dataset. }
\label{tbl:OCHumanComparison}
\vspace{-5pt}
\end{table}

\section{Additional Discussion}

\paragraph{Limitation}
While pose estimation, detection, and segmentation tasks share a common high-level perception and can help each other intuitively,
a mismatch in current task-wise objectives may make an ambiguity in the multi-task learning of them.
For example, human pose estimation should handle occluded regions while segmentation and detection only handle visible areas.
As shown in \cref{fig:limitation}, some pose joints in an occluded region make the bounding box expanded, and this eventually degrades the quantitative accuracy of object detection because the ground-truth bounding box is defined by the enclosing box of the visible segments of the instance.
Also, all pose joints are computed for any partial human instance (the rightmost column in \cref{fig:limitation}). In this case, even though the corresponding bbox and mask of the instance do not always cover the pose, the pose may still affect other tasks negatively.
Considering the visibility of pose joints may alleviate this problem.
Exploring this potential mismatch can be a clue to improve overall performance of multi-task learning with a unified architecture.



While we focus on handling the human class only in this paper, our query design is not limited to a human class. Learnable keypoints are currently trained in a supervised manner. To deal with more general classes, we can consider manual labeling or unsupervised keypoint learning as future work.

\paragraph{Potential negative social impact}

We use multiple GPUs for several days for training a model, and it induces a significant energy consumption.
While we use the public dataset (MS COCO 2017), there can be a privacy-related problem when collecting an additional dataset.
Also, our approach and its extension can be utilized in developing recognition models used in surveillance cameras, having the risk of a detrimental effect on people's privacy.

\end{document}